\documentclass[conference]{IEEEtran}
\IEEEoverridecommandlockouts
\usepackage{cite}
\usepackage{amsmath,amssymb,amsfonts}
\usepackage{algorithmic}
\usepackage{graphicx}
\usepackage{textcomp}
\usepackage{xcolor}

\usepackage{booktabs}
\usepackage{multirow}
\usepackage{pifont}
\usepackage[numbers,sort&compress]{natbib}
\usepackage{hyperref}

\makeatletter
\newcommand{\linebreakand}{%
  \end{@IEEEauthorhalign}
  \hfill\mbox{}\par
  \mbox{}\hfill\begin{@IEEEauthorhalign}
}
\makeatother

\def\BibTeX{{\rm B\kern-.05em{\sc i\kern-.025em b}\kern-.08em
    T\kern-.1667em\lower.7ex\hbox{E}\kern-.125emX}}
\begin{document}

\title{Protective Perturbations against Unauthorized Data Usage in Diffusion-based Image Generation
\thanks{The authors acknowledge the financial support of this work by the Hong Kong RGC Research Impact Fund grant with Ref. No. R1012-21.}
}

\author{\IEEEauthorblockN{1\textsuperscript{st} Sen Peng}
\IEEEauthorblockA{\textit{Department of Computer Science} \\
\textit{City University of Hong Kong}\\
Kowloon, Hong Kong SAR
}
\and
\IEEEauthorblockN{2\textsuperscript{nd} Jijia Yang}
\IEEEauthorblockA{\textit{Department of Computer Science} \\
\textit{City University of Hong Kong}\\
Kowloon, Hong Kong SAR}
\and
\IEEEauthorblockN{3\textsuperscript{rd} Mingyue Wang}
\IEEEauthorblockA{\textit{Department of Computer Science} \\
\textit{City University of Hong Kong}\\
Kowloon, Hong Kong SAR}

\linebreakand
\IEEEauthorblockN{4\textsuperscript{th} Jianfei He}
\IEEEauthorblockA{\textit{Department of Computer Science} \\
\textit{City University of Hong Kong}\\
Kowloon, Hong Kong SAR}
\and
\IEEEauthorblockN{5\textsuperscript{th} Xiaohua Jia}
\IEEEauthorblockA{\textit{Department of Computer Science} \\
\textit{City University of Hong Kong}\\
Kowloon, Hong Kong SAR}
}

\maketitle

\begin{abstract}
Diffusion-based text-to-image models have shown immense potential for various image-related tasks. 
However, despite their prominence and popularity, customizing these models using unauthorized data also brings serious privacy and intellectual property issues. 
Existing methods introduce protective perturbations based on adversarial attacks, which are applied to the customization samples.
In this systematization of knowledge, we present a comprehensive survey of protective perturbation methods designed to prevent unauthorized data usage in diffusion-based image generation.
We establish the threat model and categorize the downstream tasks relevant to these methods, providing a detailed analysis of their designs. 
We also propose a completed evaluation framework for these perturbation techniques, aiming to advance research in this field.
\end{abstract}


\section{Introduction}
Diffusion-based text-to-image models have emerged as powerful tools for a wide range of image-related tasks, including text-driven synthesis~\cite{dhariwal2021diffusion,rombach2022high} and text-driven manipulation~\cite{saharia2022palette,kawar2023imagic}.
Image samples are learned as distributions through a forward and reverse diffusion process, which can be sampled for generation with conditions. 
Latent Diffusion Models (LDMs)~\cite{rombach2022high} conduct the entire diffusion process in the latent space, improving training efficiency and inference performance. 
By leveraging open-source pre-trained LDMs like Stable Diffusion~\cite{stablediffusion} and DALLE2~\cite{ramesh2022hierarchical}, users can effectively customize models for specific downstream tasks, significantly boosting the community's development.
However, customization using unauthorized data raises significant privacy and intellectual property (IP) concerns~\cite{Dixit2023}.
Malicious users can exploit customized models to generate content that violates IP rights or even create fake images for criminal purposes. 
These adversarial customization undoubtedly bring severe ethical challenges to the community.

To mitigate this threat, existing approaches introduce adversarial noise~\cite{goodfellow2014explaining} towards the target diffusion model as protective perturbations, which are applied to the customization samples.
The applied perturbations are designed to be imperceptible yet effective in disabling the customized models when used in malicious downstream tasks.
These perturbation methods introduce various strategies to address the unique optimization problem of generating adversarial examples for diffusion models. Typically, they target two key components within LDMs: the latent autoencoder and the denoising module.
The optimization objective is to design perturbations that either shift the image's representation away from its original semantic meaning towards a target through targeted adversarial attacks, or degrade the model's performance by maximizing the denoising loss.
Several adaptive attacks~\cite{sandoval2023jpeg,cao2024impress} have been proposed targeting existing perturbation methods, including natural pre-processing transformations like JPEG compression and the addition of Gaussian noise. 
However, a systematic understanding of existing perturbation methods is still lacking.

In this paper, we provide a comprehensive survey of existing protective perturbation methods against unauthorized data usage in diffusion-based image generation. 
We start by formulating the threat model from the data owner's perspective, aligning our efforts with the community.
We then categorize the downstream tasks targeted by malicious users for customization into two main areas: 1) text-driven image synthesis, which includes object-driven synthesis and style mimicry, and 2) text-driven image manipulation, which includes image editing.
Lastly, we propose a comprehensive framework for evaluating perturbation methods, containing four key aspects: visibility, effectiveness, cost, and robustness, which can standardize and unify the evaluation criteria for protective perturbation methods.

\section{Preliminaries}
\subsection{Diffusion-based Image Generation}
The concept of diffusion-based generative models is inspired by the diffusion process in non-equilibrium thermodynamics~\cite{sohl2015deep}.
Diffusion models learn the underlying data distribution by training on a stepwise forward diffusion process.
Using a backbone model, such as UNet~\cite{ho2020denoising}, to predict reconstruction noise during the reverse diffusion process, image samples are generated through a denoising procedure.
Text prompts have been introduced to guide the image generation~\cite{nichol2021glide}, demonstrating superior performance in text-to-image synthesis~\cite{dhariwal2021diffusion} such as in DALLE 2~\cite{ramesh2022hierarchical}.
LDMs~\cite{rombach2022high} are later proposed to improve the training and inference efficiency while maintaining the generation quality through encoding the training samples into the latent space.
As an open-source large-scale LDM, Stable Diffusion (SD)~\cite{stablediffusion} is released, significantly boosting the development of diverse downstream applications in diffusion-based image generation.

\subsection{Customization of Diffusion-based Generative Models}
Diffusion-based pre-trained models can be customized for various downstream tasks by fine-tuning the personalized dataset.
Parameter-Efficient Fine-Tuning (PEFT) techniques have been proposed to improve the performance and efficiency of diffusion model customization.
DreamBooth~\cite{ruiz2023dreambooth} fine-tunes the diffusion model using only a few images to learn the customization subject, incorporating a prior preservation loss to prevent overfitting and maintain the model’s generalization capabilities.
Textual Inversion~\cite{gal2023image} optimizes a unique, non-existent trigger token for the target customization subject, which is then used to guide the image generation process.
LoRA~\cite{hu2022lora} was initially introduced to reduce the cost of fine-tuning large language models and has since been widely adopted for personalizing diffusion models.
During customization, tiny adaptation layers are injected into the frozen pre-trained model to enable updates based on the assumption that the updated model weights form a low-rank matrix, which its rank decomposition can approximate.

\subsection{Threat Model for Unauthorized Data Usage}
Customizing diffusion-based generative models using unauthorized data can severely violate the rights of data owners.
To align our effects with the community, we consider the data owner, whose data is utilized without authorization, as the defender.
The malicious user who exploits unauthorized data for customization purposes is considered the attacker.
We formulate the threat model for protective perturbations against unauthorized data usage.
To prevent attackers from exploiting unauthorized data $\mathcal{D}$ for obtaining the customized diffusion model $\mathcal{M}_{\mathcal{A}}$ for a specific downstream task, the defender applies an imperceptible noise $\delta$ as the protective perturbation to the data, resulting in the protected dataset $\mathcal{D}_{p}$.
The defender can then choose to release the protected data to the public.
When an attacker utilizes $\mathcal{D}_{p}$ for customization to obtain the model $\mathcal{M}_{\mathcal{A}}^{p}$, the performance on the target downstream task degrades significantly, resulting in failure.

\section{Formulation of Downstream Tasks}
Before delving into the specifics of current protective perturbation techniques and their corresponding adaptive attacks, it is crucial to formulate the downstream tasks for which the customization of diffusion-based text-to-image generation targets.
We initially categorize the majority of existing tasks into two primary groups: 1) text-driven image synthesis and 2) text-driven image manipulation, depending on whether the task requires an image as part of the input.

\subsection{Text-driven Image Synthesis}
Downstream tasks belonging to text-driven image synthesis exclusively require text prompts as part of the input for generation. 
Since customization in these tasks mainly focuses on learning the subject from a given personal dataset, these tasks are also referred to as \textbf{subject-driven synthesis} in some studies.
The learned subject can be a specific object, such as a person, or a concept, such as a style. 
Accordingly, we further divide text-driven image synthesis tasks into two categories: 1) object-driven synthesis and 2) style mimicry.

\subsubsection{Object-driven Synthesis}
Object-driven synthesis tasks focus on learning a specific object as the subject through the provided customization dataset.
The learned object in these tasks can be some identities with intellectual property, such as characters from a well-known cartoon.
Without authorization, the generated images of these identities can severely harm the intellectual property with minimal cost.
In some cases, this object can be the portrait of a specific individual, such as a celebrity, as detailed in~\cite{vanle2023dreambooth,wang2024simac}.
The generated photo-realistic images can help fabricate fake news, severely threatening individual reputations. 
This is regarded as one of the most serious AI-related criminal threats.

\subsubsection{Style Mimicry}
Style mimicry involves imitating a specific style based on a given customization dataset.
The learned style here serves as the subject of the downstream task.
It is commonly recognized that the style of the images is often closely associated with intellectual property, such as the distinct style of an artist's paintings. 
Considering this fact, concerns about unauthorized data usage in style mimicry are widely discussed.
Style mimicry task typically employs text prompts such as "a painting in \textit{sks} style" or "a painting by \textit{sks}" as the only input for conditioning the generation, where \textit{sks} denotes the name of the style or the artist.

\subsection{Text-driven Image Manipulation}
Text-driven image manipulation refers to downstream tasks requiring text prompts and initial images as input.
These tasks, powered by advanced generative models, aim to modify the given initial images guided by the text prompts using the learned subject.
We focus exclusively on the main category in text-driven image manipulation tasks: image editing.

\subsubsection{Image Editing}
Image editing tasks that utilize a mask require an initial image and a designated mask region as inputs. 
The text prompt is then used to guide the editing process, while the mask identifies the specific area to be edited.
In some studies, this task is also called \textbf{prompt-guided image inpainting}.
Image editing tasks can also be performed without using a mask, only relying on a text prompt to guide editing across the entire region of the initial image.
If the learned subject is a style, the image editing task becomes a style transfer task.
Given an initial image and a target style described in the text prompt, the model can generate a transformed image that follows the target style.

\section{Protective Perturbations}
\begin{table*}[htbp]
\caption{Overview of the protective perturbation methods against unauthorized data usage in diffusion-based image generation.}
\begin{center}
\begin{tabular}{@{}l|c|cc|ccc@{}}
\toprule
\multicolumn{1}{c|}{\multirow{2}{*}{Perturbation}} & \multirow{2}{*}{Year} & \multicolumn{2}{c|}{Optimization Objective} & \multicolumn{3}{c}{Downstream Task} \\ \cmidrule(l){3-7} 
\multicolumn{1}{c|}{} &  & Latent AutoEncoder & E2E Diffusion Process & Object-driven Sythesis & Style Mimicry & Image Editing \\ \midrule
AdvDM~\cite{liang2023adversarial} & \multirow{11}{*}{2023} & \ding{55} & \ding{51} & \ding{51} & \ding{51} & \ding{51} \\
Photoguard~\cite{salman2023raising} &  & \ding{51} & \ding{51} & \ding{55} & \ding{55} & \ding{51} \\
Glaze~\cite{shan2023glaze} &  & \ding{51} & \ding{55} & \ding{55} & \ding{51} & \ding{55} \\
Anti-DB~\cite{vanle2023dreambooth} &  & \ding{55} & \ding{51} & \ding{51} & \ding{51} & \ding{55} \\
Mist~\cite{liang2023mist} &  & \ding{51} & \ding{51} & \ding{51} & \ding{51} & \ding{51} \\
EUDP~\cite{zhao2024unlearnable} &  & \ding{55} & \ding{51} & \ding{51} & \ding{51} & \ding{55} \\
Zhang et al.~\cite{zhang2023robustness} &  & \ding{51} & \ding{51} & \ding{55} & \ding{55} & \ding{51} \\
SDS~\cite{xue2024effective} &  & \ding{51} & \ding{51} & \ding{51} & \ding{51} & \ding{51} \\
Nightshade~\cite{shan2024nightshade} &  & \ding{51} & \ding{55} & \ding{51} & \ding{51} & \ding{55} \\
MetaCloak~\cite{liu2024metacloak} &  & \ding{55} & \ding{51} & \ding{51} & \ding{51} & \ding{55} \\
SimAC~\cite{wang2024simac} &  & \ding{55} & \ding{51} & \ding{51} & \ding{51} & \ding{55} \\ \midrule
Chen et al.~\cite{chen2024exploring} & \multirow{5}{*}{2024} & \ding{51} & \ding{51} & \ding{51} & \ding{51} & \ding{51} \\
ACE~\cite{zheng2024improving} &  & \ding{51} & \ding{51} & \ding{51} & \ding{51} & \ding{51} \\
VCPro~\cite{mi2024visual} &  & \ding{55} & \ding{51} & \ding{51} & \ding{51} & \ding{55} \\
PCA~\cite{guo2024grey} &  & \ding{55} & \ding{51} & \ding{55} & \ding{55} & \ding{51} \\
PAP~\cite{wan2024prompt} &  & \ding{55} & \ding{51} & \ding{51} & \ding{51} & \ding{55} \\ \bottomrule
\end{tabular}
\label{tab1}
\end{center}
\end{table*}

\subsection{Existing Methods}
To the best of our knowledge, the earliest study on protective perturbations is AdvDM~\cite{liang2023adversarial}. 
It is proposed to defend against malicious customization in text-driven image synthesis tasks, specifically those leveraging Textual Inversion. 
AdvDM can also prevent unauthorized data usage in text-driven image manipulation tasks, including style transfer and image-to-image synthesis.
It first introduces the idea of using adversarial noise against the target diffusion model as the protective perturbation, preventing the model from learning the perturbed samples.
Adversarial samples are generated by maximizing the original loss function during the diffusion process training.
By optimizing over the entire time sequence of the diffusion process using the Monte Carlo method, AdvDM generates perturbations that prevent the model from learning the target subject's representation, which cannot be used as the condition in customization.
The optimization targets the diffusion process within the LDM. 
Additionally, the authors evaluate the robustness of AdvDM against adaptive attacks, including natural transformation techniques.

Concurrently, Photoguard~\cite{salman2023raising} is introduced to prevent unauthorized data leakage from malicious image editing, which also employs adversarial noise as the protective perturbation.
For LDMs, Photoguard utilizes two protection strategies: targeting the latent encoder and targeting the diffusion process.
Unlike AdvDM, Photoguard employs targeted adversarial attacks in both of its strategies.
Specifically, the encoder attack generates the perturbation by minimizing the difference between the perturbed image and a targeted image, typically as pure noise or a random pattern.
The diffusion attack generates the perturbation in an end-to-end way by minimizing the difference between a target image and the output of the diffusion model, using the initial image as input.
Due to the high memory requirements of optimizing perturbations, Photoguard only performs the back-propagation through a few timesteps in the diffusion process.
The authors also demonstrate that perturbations generated by Photoguard exhibit weak robustness. 
Therefore, it is essential to establish policies to mitigate malicious image editing using AI.

Glaze~\cite{shan2023glaze} is designed to defend against unauthorized data usage, specifically in the downstream task of style mimicry.
The authors point out that Photoguard is not explicitly designed to prevent style mimicry, which results in a shift of all features in the protected images.
Instead, Glaze generates perturbations that specifically alter features related to the style of the customized images.
To achieve this, Glaze first utilizes an existing pre-trained model to perform style transfer on customization dataset images.
The transferred images are unrelated to the subject's original style.
Then, perturbations are generated by minimizing the distance between the latent features of the sample image and those of the corresponding target image.
Glaze employs the Learned Perceptual Image Patch Similarity (LPIPS) loss, rather than the L$p$ norm, to constrain the constructed perturbation during optimization.

Anti-DreamBooth (Anti-DB)~\cite{vanle2023dreambooth} focuses on the defense against unauthorized customization using DreamBooth, which targets text-driven synthesis as its downstream task.
DreamBooth requires only a small number of samples for customization, as it fine-tunes the model by minimizing the diffusion training loss alongside a prior preservation loss. 
This approach helps maintain the generalization of the customized model.
Anti-DB first introduces the Fully-trained Surrogate Model Guidance (FSMG) strategy, which induces overfitting in the customized model by maximizing the diffusion training loss on clean samples while minimizing the prior preservation loss on customization samples.
Additionally, the authors highlight that, unlike in the methods mentioned above, where the target surrogate model is typically assumed to be fixed, the target surrogate model used in DreamBooth should not be static.
To address this issue, Anti-DB introduces Alternating Surrogate and Perturbation Learning (ASPL), which optimizes by iteratively updating the perturbations and surrogate model weights.
Both strategies can be combined with targeted adversarial attack approaches. 

Mist~\cite{liang2023mist} is a follow-up method to AdvDM, designed to enhance protective perturbations. 
It introduces two types of adversarial losses for constructing the perturbations: 1) semantic loss, which targets the diffusion process, and 2) textual loss, which focuses on the latent encoder utilized by LDMs.
To improve the effectiveness of protective perturbations, Mist combines these two loss functions during optimization, using a balancing factor. Additionally, the authors highlight the importance of carefully selecting the target image for latent space attacks, as the performance of attacking the textual loss is susceptible to this choice. 
It is recommended that images with a high contrast ratio and sharp edges should be selected as the target image for optimal results.

The authors in~\cite{zhao2024unlearnable} propose adding imperceptible adversarial noise as protective perturbations on customization samples to make them unlearnable for diffusion-based image generative models.
They observe that as the diffusion timestep becomes larger, the effect of the protective noise decreases, while the effect of the noise introduced during the forward diffusion process increases.
Therefore, they propose Enhanced Unlearnable Diffusion Perturbation (EUDP), which integrates a timestep sampling method based on their observations.
EUDP also targets the diffusion process, aiming to select the more critical timesteps during optimization to construct effective perturbations.
Experimental results demonstrate that EUDP outperforms AdvDM and can be effectively applied to LDMs.

In~\cite{zhang2023robustness}, the authors propose generating perturbations via adversarial attacks targeting different modules within LDMs. 
Additionally, they discuss the underlying assumptions in the adversarial attack settings for the malicious image editing downstream task.
They define the white-box setting as having full access to the target model's information, while the black-box setting assumes no access to such information. 
This assumption can also be discussed when evaluating the transferability of protective perturbations.

In \cite{xue2024effective}, the authors identify the latent encoder as the bottleneck for adversarial attacks targeting the end-to-end diffusion process of LDMs. 
They show that directly attacking the diffusion process in the latent space is ineffective, even when a larger perturbation budget is applied.
This phenomenon demonstrates the robustness of diffusion denoisers against adversarial samples. 
Furthermore, they observe that, instead of using gradient ascent for optimization, gradient descent can generate even more imperceptible perturbations.
To this end, they propose Score Distillation Sampling (SDS), which leverages score distillation to accelerate semantic loss computation. 
SDS can also be combined with textual loss, a method called SDST. 
The concept of using gradient descent for optimization is also applicable in AdvDM.
Experimental results demonstrate that SDS can generate more invisible perturbations with lower computational cost.

Unlike the methods discussed above, Nightshade~\cite{shan2024nightshade} introduces an optimized prompt-specific poisoning attack to generate protective perturbations. 
The authors highlight that the effectiveness of perturbation techniques like Glaze is limited, often requiring over 800 samples to protect a single concept.
Instead, Nightshade constructs perturbation, which can shift the representations of customization samples away from their semantic meaning, pushing them toward an irrelevant concept in the latent space.
It first selects an irrelevant concept distinct from the one in the customization data and then generates auxiliary images based on this unrelated concept. Then, it minimizes the difference between the perturbed sample and a randomly selected image from the auxiliary set, shifting its representation.
Customization on these samples can prevent the model from learning the intended concept and even lead to a collapse in its original task performance.

In~\cite{liu2024metacloak}, the authors highlight that protective perturbations against DreamBooth differ in that they adapt to the model’s active learning during the customization process.
They argue that Anti-DB employs a challenging bi-level optimization process based on hand-crafted heuristics, which leads to sub-optimal performance. Additionally, existing perturbation methods lack robustness against fundamental natural transformations.
To address these challenges, they propose MetaCloak, which enhances the Anti-DB method by optimizing using transformed perturbed images.
By incorporating some natural transformations into the optimization process, MetaCloak can produce more robust perturbations.
Moreover, it also employs an iterative approach to optimize the perturbation by looking ahead to K-steps in each iteration. 
MetaCloak is a transformation-robust, model-agnostic attack, as the iterative process dynamically incorporates target surrogate models.

Similar to the findings in~\cite{zhang2023robustness}, the authors in~\cite{wang2024simac} also observe that the timesteps involved in optimizing protective perturbations have varying effects.
They find that lower timesteps contribute more to the effectiveness of perturbations. 
Based on this insight, they propose the SimAC method.
This adaptive approach to existing perturbation methods selects an optimal time interval to target during optimization. 
Moreover, it combines a feature loss with the vanilla training loss for overall optimization, which calculates layer-wise distances between features in specifically selected layers of the denoising model (UNet). 
Experimental results demonstrate that adapting SimAC to Anti-DB can improve its performance.

Additionally, the authors in~\cite{chen2024exploring} also find that the importance of timesteps in optimizing protective perturbations varies.
They treat the model at different timesteps as distinct surrogate models for adversarial attacks, following the Monte Carlo sampling approach.
They observe that the smoothness of surrogate models varies across different timesteps. 
As a result, they only select optimal time intervals for existing methods like AdvDM and Mist. 
Experiments demonstrate that adapting specific time interval selections can improve the performance.

Following Mist, the authors in~\cite{zheng2024improving} introduce Attacking with Consistent Score-Function Errors (ACE). 
They reveal the dynamics described in~\cite{song2021score} when learning perturbed samples in LDMs. 
Their results demonstrate that adversarial noise introduces an additional error term in score function prediction, which can be mitigated by incorporating a learned bias during the reverse diffusion process.
Moreover, they observe that the pattern of reverse bias varies across different images. 
Therefore, they propose unifying the reverse bias pattern during perturbation construction. 
This consistent pattern can then be set as a specific target image.

The authors in~\cite{mi2024visual} point out that the above methods produce perturbations that are not entirely "imperceptible."
In fact, these perturbations remain visible and can even negatively impact certain downstream tasks.
The reason is that these methods generate perturbations across the entire region of the target image.
Recognizing the need for a more subtle and stealthy adversarial noise, the authors propose VCPro, a method that prioritizes protecting the essential information in the target image.
Users can first apply models like SAM~\cite{kirillov2023segment} to segment the target image and extract the subject's identity region. 
A user-protective mask is then generated and utilized in a regional adversarial learning loss. 
Experimental results demonstrate that VCPro ensures low visibility of perturbations while maintaining high effectiveness.

In~\cite{guo2024grey}, the authors argue that existing perturbation methods rely on the white-box assumption, which is an unrealistic scenario that grants full access to the target model's information. 
Furthermore, they observe that these perturbations can only partially disable the customization model in downstream tasks, which tend to introduce chaotic patterns rather than fully degrading the semantic quality of the manipulated outputs.
To address these challenges, they propose the Posterior Collapse Attack (PCA), which leverages the observation that the VAEs used in LDMs are vulnerable to posterior collapse during training.
PCA introduces a posterior collapse loss during the optimization by minimizing the KL divergence between the target distribution and the learned posterior distribution.
Experimental results demonstrate its superior effectiveness in countering malicious image editing tasks.

In~\cite{wan2024prompt}, the authors observe that existing perturbations are typically generated using specific training prompts tailored to individual target images. 
However, these perturbations can become ineffective if an attacker employs different prompts during inference.
To tackle this issue, the authors propose Prompt-Agnostic Adversarial Perturbation (PAP) based on samples drawn from a prompt distribution modeled using Laplace approximation rather than relying on specific prompts.

\subsection{Evaluation Metrics}
Evaluation of the proposed perturbation methods involves the following four aspects:

\subsubsection{Perturbation Visibility}
The perturbation visibility metric measures the visibility of added protective perturbations, which the degradation of image quality can also represent after applying protective measures. 
In existing studies, the visibility of perturbations is typically evaluated by comparing the perturbed images to the original ones.
Many existing metrics can be used, including the Structural Similarity Index Measure (SSIM)~\cite{wang2004image}, Peak Signal-to-Noise Ratio (PSNR)~\cite{huffman1952method}, Learned Perceptual Image Patch Similarity (LPIPS), and Fréchet Inception Distance (FID)~\cite{heusel2017gans}. 
A higher similarity between the perturbed and original images results in better imperceptibility of the constructed protective perturbations.

\subsubsection{Perturbation Effectiveness}
Perturbation effectiveness measures how protective perturbations degrade performance in malicious downstream tasks that use unauthorized data.
Since the effectiveness is related to the specific downstream task the perturbations target, evaluation requires comparing the performance of the customized model $\mathcal{M}(D)$ trained on clean customization data $D$ with that of $\mathcal{M}(D_{p})$, trained on protected perturbed data $D_{p}$.
We compute the perturbation effectiveness score for the target downstream task $\mathcal{T}$ as follows:
\begin{equation}
    S_{\text{PE}}(f(D)) = Q^{\mathcal{T}}_{\text{PE}}(\mathcal{M}(D)) - Q^{\mathcal{T}}_{\text{PE}}(\mathcal{M}(D_{p})),
\end{equation}
where $f$ is the perturbation method that has $f(D)=D_{p}$ and $Q^{\mathcal{T}}$ represents the evaluation metric for task $\mathcal{T}$. 
A larger performance degradation indicates a higher effectiveness of the constructed perturbations.

\subsubsection{Perturbation Cost}
The perturbation cost quantifies the computational overhead associated with generating perturbations. 
It is typically evaluated using metrics such as time or memory consumption while constructing the perturbations.

\subsubsection{Perturbation Robustness}
Perturbation robustness is evaluated by measuring the degradation in a perturbation method's effectiveness score after applying adaptive attacks. 
Several studies have shown that adaptive attacks can significantly reduce the effectiveness of perturbations~\cite{sandoval2023jpeg,cao2024impress}."
The perturbation robustness score is computed as follows:
\begin{equation}
    S_{\text{PR}}(f(D)) = S_{\text{PE}}(f(D)) - S_{\text{PE}}(\mathcal{A}(f(D))),
\end{equation}
where $\mathcal{A}$ represents the adaptive attack function applied to the perturbed samples.

\section{Conclusions and Future Works}
This paper presents a comprehensive survey of existing protective perturbation methods. 
We formulate the threat model and downstream tasks for these methods, presenting a detailed analysis of their designs. 
Additionally, we propose a complete framework for evaluating these perturbation techniques. 
Future research may focus on enhancing the robustness of perturbation methods against existing and potential adaptive attacks.
We believe this systematization of knowledge will help the research community advance in this field.

\bibliographystyle{IEEEtran}
\bibliography{ref}

\end{document}